\documentclass[conference]{IEEEtran}
\IEEEoverridecommandlockouts

\usepackage{cite}
\usepackage{amsmath,amssymb,amsfonts}
\usepackage{algorithmic}
\usepackage{graphicx}
\usepackage{textcomp}
\usepackage{xcolor}
\usepackage{booktabs}
\usepackage{tabularx}
\usepackage{array}
\usepackage{multirow}
\usepackage{makecell}

\def\BibTeX{{\rm B\kern-.05em{\sc i\kern-.025em b}\kern-.08em
    T\kern-.1667em\lower.7ex\hbox{E}\kern-.125emX}}
\begin{document}

\title{MARCUS: Missing-Aware Region Representation with Contextual Urban Signals for Rent Prediction}

\author{
\IEEEauthorblockN{
Chenya Huang\textsuperscript{1},
Bin Liang\textsuperscript{1,*},
Zhidong Li\textsuperscript{1},
Yuxi Lu\textsuperscript{1},
Kunqi Li\textsuperscript{1},
Justin Wang\textsuperscript{2},
and Fang Chen\textsuperscript{1}
}

\IEEEauthorblockA{
\textsuperscript{1}\textit{Data Science Institute}\quad
\textit{University of Technology Sydney}\quad
Sydney, Australia\\
\{Bin.Liang, Zhidong.Li, Fang.Chen\}@uts.edu.au\quad
\{Chenya.Huang, Yuxi.Lu, Kunqi.Li\}@student.uts.edu.au\\
\textsuperscript{*}Corresponding author: Bin Liang.
}

\IEEEauthorblockA{
\textsuperscript{2}\textit{The Property Investors Alliance}\quad
Sydney, Australia\quad
Justin@pia.com.au
}
}

\maketitle

\begin{abstract}
Multimodal urban data has expanded the applications of urban region representation learning, such as functional zone identification and real estate appraisal, but also introduces challenges caused by data incompleteness. Existing studies usually handle missing data through imputation, treating missingness as noise while ignoring its potential semantic value. To address this issue, we propose MARCUS, a missing-aware region representation model that treats missingness as a contextual urban signal. MARCUS models missingness in three stages: Intra Learning jointly encodes observed features and missing patterns, Inter Learning estimates modality reliability to guide cross-modal interaction, and Fusion uses missing-aware and time-aware gating to generate the final region embedding. We apply MARCUS to rent prediction, a task with long-term trends and seasonal fluctuations, using real-world datasets from Sydney and New York. Experimental results show that MARCUS achieves state-of-the-art performance, reducing MAE by 51.35\% on Sydney and 12.62\% on New York compared with the best baselines. Additional experiments, including an imputation-based ablation study and randomized additional-missingness analysis, further demonstrate the effectiveness of the proposed method. For reproducibility, we release the implementation details and source code at: https://github.com/Santooops/MARCUS

\end{abstract}


\section{Introduction}
In recent years, urban computing has attracted increasing attention from the research community~\cite{b1}. This trend is driven by the growing availability of multimodal urban data~\cite{b2}. In urban computing, cities are commonly partitioned into multiple regions with well-defined boundaries but irregular shapes~\cite{b3}. These regions differ substantially in functional attributes, demographic structure, and economic activities. As a result, effectively modelling and understanding urban structure has become a key problem. High-quality urban region representations help characterize the spatial structure and functional distribution of cities. They also support a variety of downstream tasks, such as housing price prediction \cite{b4}, \cite{b5}, rent estimation \cite{b6}, crime prediction \cite{b7}, \cite{b8}, and urban vitality assessment~\cite{b9}. 

\begin{figure}[t]
    \centering
    \includegraphics[width=1.03\linewidth]{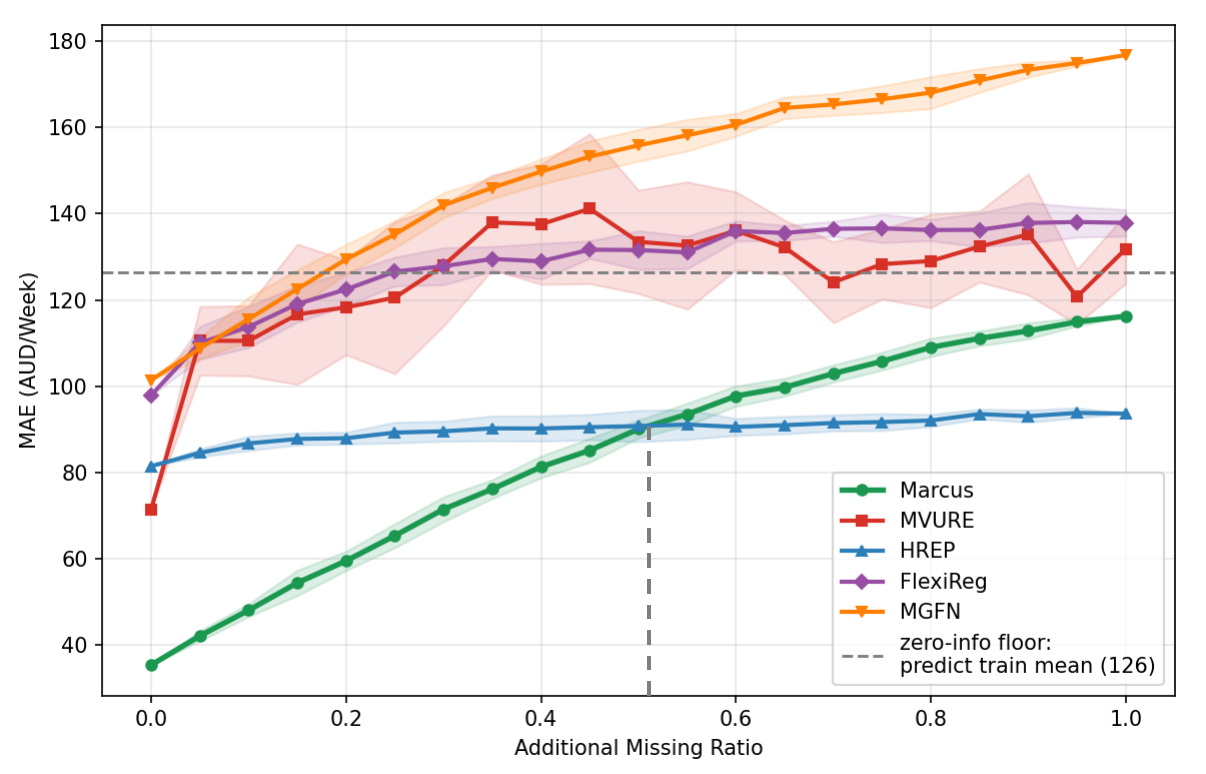}
    \caption{Additional missing ratio analysis on the Sydney dataset. The additional missing ratio denotes the proportion of originally observed data that are artificially masked. Naturally missing records in the original data are kept unchanged.}
    \label{fig:ratio}
\end{figure}

However, multimodal urban data is often incomplete in real urban environments. Coverage across multiple modalities typically exhibits systematic missingness. Some modalities may be entirely unavailable in certain regions. Others are only partially observed in space or time. More importantly, such missingness is usually not random noise. It is highly correlated with regional attributes and urban structure. For example, population-related indicators are semantically invalid in uninhabited areas, and POI or mobility signals tend to be sparser in newly developed or more remote regions. The resulting missingness patterns therefore carry urban semantics. They reflect functional status and development disparities across regions. Hence, missingness should be treated as an urban context signal, rather than merely a data issue to be resolved \cite{b10}.

Existing methods\cite{b6}, \cite{b39}, \cite{b40} often treat missingness as an imputation problem, or adopt simple masking strategies during fusion. However, these approaches suffer from two key limitations.

\textbf{1)} They make it difficult for the model to explicitly distinguish reliable information from signals originating from sparse or missing sources during cross-modal interaction and fusion. This can introduce bias and noise.

\textbf{2)} They discard the regional semantics embedded in missingness patterns, preventing such information from contributing to representation learning. Even with dedicated imputation models, additional assumptions and imputation errors are inevitable, which may further amplify bias.

To address these limitations, we propose a missing-aware urban region representation learning model. It explicitly characterizes availability patterns in multimodal urban data and allows missingness information to permeate key stages of representation learning, especially during cross-modal interaction and fusion. Specifically, in intra learning, we jointly model observed-value features and missingness features by projecting them into the same latent space, then fusing the two embeddings into a unified representation. In inter learning, we estimate modality reliability based on missing status, thereby adaptively modulating interaction strength. In the final fusion stage, we further incorporate missingness patterns and temporal context to perform gated weighting, yielding region representations that are more robust and semantically informative under incomplete multimodal urban data. In addition, to evaluate model performance under different degrees of missingness, we conduct an additional missing ratio experiment, as shown in Fig.~\ref{fig:ratio}. MARCUS consistently achieves the best performance when the additional missing ratio is below approximately 55\%. More detailed analysis is provided in Section ``Analysis on Missing Ratio''. 

Our contributions can be summarized as follows:

\begin{itemize}
\item We propose a technical formulation that treats multimodal missingness as an informative urban context signal. For each region, time step, and modality, we construct missingness descriptors containing availability, missing ratio, and missing-location information, and inject them into intra learning, inter learning, and final modality fusion. This allows MARCUS to adaptively assess modality reliability and learn robust region representations from incomplete multimodal urban data.

\item We propose a missingness-driven region representation learning framework, where missingness information permeates the entire pipeline, from intra-modality encoding to cross-modality interaction and final fusion.

\item We validate MARCUS on real-world rental price prediction using datasets from Sydney and New York. MARCUS consistently outperforms traditional predictors and region embedding baselines across all metrics. Further comparisons with an imputation-first strategy and additional missingness experiments (Fig.~\ref{fig:ratio}) show that explicitly modelling missingness provides useful predictive information and enables MARCUS to remain robust under severe modality incompleteness.

%
\end{itemize}

\section{Related Work}

Early studies on urban region representation learning primarily relied on human mobility data to construct region embeddings, as mobility patterns capture dynamic interactions and behavioral connections between regions~\cite{b2,b22,b23,b18,b24}. However, mobility data alone reflects only one aspect of urban dynamics and may overlook the semantic, functional, and socioeconomic characteristics of urban regions.


To obtain a more comprehensive representation of regions, recent studies have begun to jointly model mobility data with region attributes \cite{b25}, \cite{b26}, \cite{b27}, \cite{b23}. For instance, POI distributions provide cues about regional functions \cite{b28}, \cite{b29}, \cite{b30}, \cite{b31}. Census statistics capture demographic structure and socioeconomic context. Street-view or remote-sensing imagery further describes the built environment and land-use patterns  \cite{b25}, \cite{b32}, \cite{b20}, \cite{b33}, \cite{b34}, \cite{b35}. In addition, text, as a form of human language, also contributes to the enhancement of urban embedding \cite{b29}, \cite{b20}, \cite{b36}. These heterogeneous sources provide complementary perspectives on urban regions and have motivated multi-source region embedding learning. In parallel, researchers construct inter-region graphs based on spatial adjacency, functional similarity, or mobility \cite{b37}, \cite{b38}, \cite{b23}, \cite{b18}, \cite{b17}.
These methods help capture cross-region dependencies and improve the structural consistency of the learned representations.

However, existing fusion and relational modelling methods often rely on an implicit assumption. It is that most modalities are available across regions with comparable quality, enabling alignment and joint learning on a unified spatial unit. When this assumption does not hold, missingness and noise can be amplified during fusion and message passing. This should bias region representations and undermine structural consistency. Motivated by this practical constraint, we propose a novel missing-aware multimodal urban region representation learning framework to address this issue. For heterogeneous data, we characterize data availability through missingness patterns and use them to guide cross-modal fusion weights, resulting in more robust and transferable urban region representations.

In addition, some studies have introduced prompt-based task adaptation into urban region representation learning.HREP adopts a prefix-tuning–style prompt to guide heterogeneous region graph representations to learn more effectively across tasks \cite{b19}. FlexiReg further leverages prompting to customize task-oriented region representations \cite{b20}. These studies show that prompt-based adaptation can improve the flexibility and transferability of region embeddings. However, they mainly focus on task adaptation and do not explicitly model multimodal missingness, which is the focus of this work.

\section{Data Description and Analysis}

\begin{table}
  \centering
  \caption{Dataset Statistics}
  \label{tab:dataset_stats}
  \begin{tabular}{l l c c}
    \toprule
    \multicolumn{1}{c}{Datasets} &
    \multicolumn{1}{c}{Description} &
    \multicolumn{1}{c}{Sydney} &
    \multicolumn{1}{c}{New York} \\
    \midrule
    Regions &  Partition scheme & SA2 & ZIP code \\
            & \# Regions        & 373 & 177 \\             
    \midrule
    Census data & \# Surveys    & 19,750  & 85,154 \\
                &  Missing ratio  & 0.54\%  & 0\%  \\
    \midrule
    POI data    & \# Records      & 145,548 & 20,570 \\
                & \# Categories   & 13 & 13 \\
                & Missing ratio  & 27.67\%  & 19.66\%  \\
    \midrule
    Traffic data & \# Mobility records & 22,816 & 240,558,409 \\
                 & Missing ratio  & 40.40\%  & 6.87\%  \\
    \midrule
    Rent data   & \# Records      & 608,652 & 6,372 \\
    \midrule
    Temporal data & Time Period &
                  \makecell{07/2021\\--\\07/2024} &
                  \makecell{01/2022\\--\\12/2024} \\
    \bottomrule
  \end{tabular}
\end{table}

In this section, we introduce the datasets used in our research and provide a preliminary analysis. We use two real-world urban datasets, Sydney and New York, which correspond to major real-estate markets in Australia and the United States, respectively. We collected Point of Interest(POI) data, census statistics, and traffic records for both cities to construct region representations. These data were obtained from the Australian Bureau of Statistics (ABS) \cite{b11}, NYC Open Data \cite{b12} and Office of Policy Development and Research \cite{b13}. In addition, we collected rental datasets for the downstream task. Table 1 summarizes the dataset statistics, including missing status. The missing ratio is computed at the region level.


\subsection{Region Data}
Due to differences in the availability and usability of other data sources across geographic scales, we adopt two different spatial partition schemes for Sydney and New York. Sydney follows the ABS statistical geography, Statistical Area Level 2 (SA2), while New York uses ZIP codes as the regional units. This setting also allows us to examine how our model performs under different spatial granularities.

\subsection{Census Data}
The raw census datasets contain 12,176 and 85,154 records for Sydney and New York. Both are reported at the finest available statistical units. To align with other data sources, we aggregate Sydney’s census data from Statistical Area Level(SA1) to SA2, and New York’s census data from Census Tracts to ZIP codes. Each record includes key socioeconomic attributes such as total population, average household income, educational attainment, and employment rate. In the Sydney census data, there is only a very small amount of missingness (0.54\%), whereas the New York census data is complete.

\begin{figure}[t]
    \centering
    \includegraphics[width=1\linewidth]{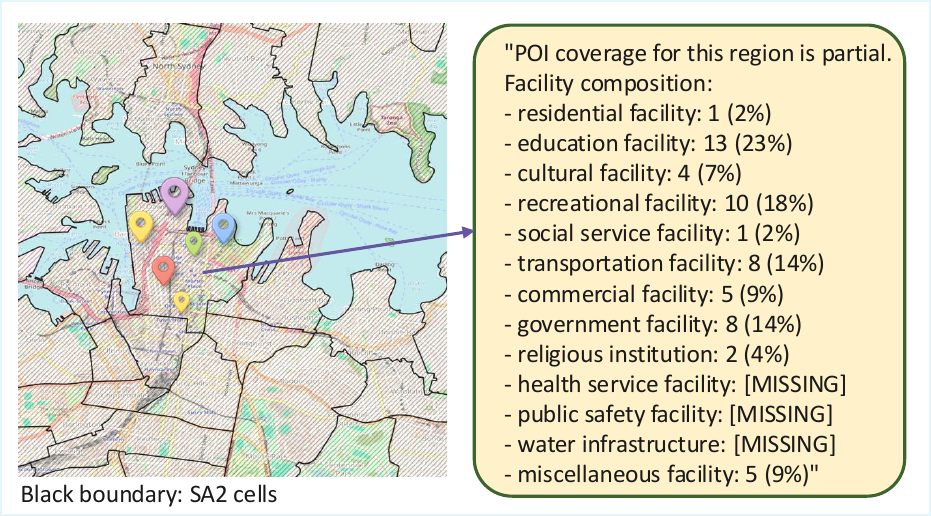}
    \caption{Example of POIs in the Sydney Region}
    \label{fig:poi}
\end{figure}

\subsection{POI Data}
We collect all POIs in Sydney and New York, which are officially categorized into 13 functional groups in each city. Although the category definitions differ slightly between the two datasets, they broadly cover common urban functions such as community, education, health, and recreational facilities. In total, there are 145,548 POIs in Sydney and 20,570 POIs in New York. For each region, we construct a textual description of the POIs it contains, as illustrated in Fig.~\ref{fig:poi}. In the POI data, Sydney has a notable amount of missingness (27.67\%), while New York has a lower missing ratio (19.66\%).

\subsection{Traffic Data}
Traffic data characterize city mobility flows. Due to data availability, we use public transport station entry and exit volumes for Sydney, and taxi trip records for New York. Taxi trip records include both street-hail and e-hail (app-based) trips, covering the entirety of New York City. These mobility data include information such as origins and destinations, and reflect the travel mode (e.g., train, metro and taxi). Since SA2 is a community-level geography designed for statistical reporting, transport stations do not cover all regions. As a result, the Sydney traffic dataset contains a substantial amount of missingness. The missing ratio of the New York traffic data is 6.87\%. A large portion of this missingness arises because the raw records contain only origins or only destinations.

\subsection{Rent Data}
\begin{figure*}
    \centering
    \includegraphics[width=1\linewidth]{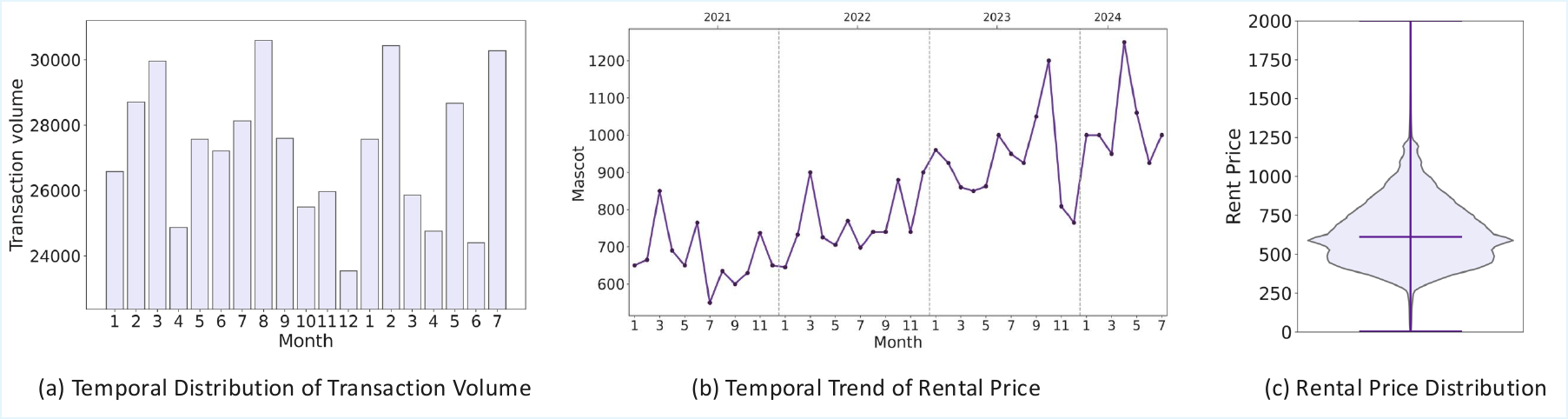}
    \caption{Analysis and Visualization on Sydney Rent Data}
    \label{fig:rent}
\end{figure*}

In this paper, rent data are collected from real estate agencies. Sydney rent data are sourced from PriceFinder \cite{b14} and the Property Industry Alliance (PIA) \cite{b15}, while the New York rent data are obtained from Zillow \cite{b16}. The New York rent data are provided as ZIP-code–level rent index, rather than transaction-level rental records. The scope of the Sydney dataset spans from July 2021 to July 2024, while the New York dataset covers January 2022 to December 2024. Fig.~\ref{fig:rent}(a) shows the monthly transaction volume from January 2023 to July 2024, revealing pronounced seasonal peaks. Fig.~\ref{fig:rent}(b) plots the rental price trend over 43 months for Mascot, a representative area in Sydney, where we observe both seasonal fluctuations and a clear long-term upward trend. For a more clear presentation, we use data spanning January 2021 to July 2024. Fig.~\ref{fig:rent}(c) presents the overall rent distribution in Sydney. 


\section{Proposed Model}

We start with a few basic concepts and a problem statement. The frequently used symbols are summarized in Table 2.

\begin{table}[htbp]
  \caption{Frequently Used Symbols}
  \label{tab:math}
  \centering
  \small
  \setlength{\tabcolsep}{4pt}
  \renewcommand{\arraystretch}{1.10}
  \begin{tabularx}{\columnwidth}{@{}l>{\raggedright\arraybackslash}X@{}}
    \toprule
    \textbf{Symbol} & \textbf{Description} \\
    \midrule
    $\mathcal{R}$ & Set of regions (e.g., SA2) \\
    $N$ & Number of regions \\
    $\mathcal{M}=\{P, C, F\}$ & Modality set (POI, census, traffic) \\
    $\mathbf{s}$ & Missingness vector for modality $m$ at $(i,t)$ (flag, ratio, and missing-location encoding). \\
    $\mathbf{h}$ & Intra-modality embedding for modality $m$ at $(i,t)$ \\
    $\mathbf{H}$ & Inter-updated embedding after cross-modal interaction \\
    $\mathbf{Z}$ & Final fused region embedding for downstream prediction \\
    $y_{i,t+1}$ & Ground-truth rent at $t{+}1$ \\
    $\hat{y}_{i,t+1}$ & Predicted rent at $t{+}1$ \\
    \bottomrule
  \end{tabularx}
\end{table}


\subsection{Features description}

\noindent\textbf{Regions.} The set of regions $\mathcal{R}$ is obtained by partitioning the spatial area of interest into non-overlapping units using a predefined scheme (e.g., postcode-based areas or census-defined regions). We denote the $i$-th region by $r_i$. For each region, we consider three types of features, including two static modalities (POI, census) and one dynamic modality (traffic), and collect them into the set $\mathcal{M}$.

\noindent\textbf{POI.} For each region, we collect all points of interest (POIs) from the Australian Bureau of Statistics and NYC Open Data, and follow the official POI taxonomies provided by the corresponding sources.
Both cities contain 13 POI categories (e.g., education and transportation), although the category systems are city-specific and not required to be aligned across cities.
For each region $r_i$, we first count the number of POIs in each category, and then serialize these statistics into a compact textual representation in the form of category count pairs (e.g., \texttt{education:12; transport:3}).
We encode the resulting text into a fixed-dimensional semantic embedding, which serves as the POI modality input to our model.

\noindent\textbf{Census.}
Consistent with the POI modality, we collect the most recent census data for both cities and extract region-level socioeconomic features, such as household income, employment rate, and educational attainment.
For each region $r_i$, we represent the census attributes as a feature vector $\mathbf{C}_i$.

\noindent\textbf{Traffic (Flow).}
For each region $r_i$, we construct month-level inter-region mobility information. For each month $t$, we define the flow $F_{i,j,t}$ for a region pair $(r_i,r_j)$, representing the travel volume from $r_i$ to $r_j$ during month $t$ (which can be summarized as inflow/outflow statistics).
As the most readily available mobility data differ across cities, we use train station tap-in/tap-out counts as a proxy for travel intensity in Sydney, and taxi trip volumes to characterize inter-region mobility in New York City.

\noindent\textbf{Missingness.}

We treat missingness as a contextual urban signal, and construct a missingness feature vector $\mathbf{s}^{(m)}_{i,t}$ for each region $r_i$, month $t$, and modality $m\in\mathcal{M}$.
Specifically, $\mathbf{s}^{(m)}_{i,t}$ includes an availability indicator $b^{(m)}_{i,t}$, a missingness ratio $\rho^{(m)}_{i,t}$, and a missing-location encoding $\mathbf{l}^{(m)}_{i,t}$ that captures where missingness occurs and its spatial pattern. 
Here, $b^{(m)}_{i,t}\in\{0,1\}$ is defined at the modality level: $b^{(m)}_{i,t}=1$ if any feature in modality $m$ is missing for region $i$ at time $t$, and $b^{(m)}_{i,t}=0$ otherwise. The missingness ratio $\rho^{(m)}_{i,t}$ is defined as the fraction of missing features in modality $m$ for region $i$ at time $t$:
\begin{equation}
\rho^{(m)}_{i,t}=\frac{\left|\mathcal{F}^{(m)}_{\mathrm{miss}}(i,t)\right|}{\left|\mathcal{F}^{(m)}(i,t)\right|}.
\end{equation}
We incorporate $\mathbf{s}^{(m)}_{i,t}$ into representation learning and cross-modal fusion, enabling the model to adaptively adjust modality contributions under incomplete observations.
\begin{equation}
\mathbf{s}_{i,t}^{(m)}=\big[\, b_{i,t}^{(m)} \,\Vert\, \rho_{i,t}^{(m)} \,\Vert\, \mathbf{l}_{i,t}^{(m)} \,\big].
\end{equation}

\subsection{Problem statement}
Given a set of regions $\mathcal{R}=\{r_i\}_{i=1}^{N}$ and a sequence of months indexed by $t$,
we observe multi-modal region attributes from three modalities $\mathcal{M}=\{P,C,F\}$, where $P$, $C$, and $F$ denote POI, census, and traffic (flow), respectively.
For each region $r_i$ at month $t$, let $\mathbf{x}^{(m)}_{i,t}$ denote the input of modality $m\in\mathcal{M}$, and let $\mathbf{s}^{(m)}_{i,t}$ denote its missingness vector (including missing flag, missing ratio, and a missing-location encoding).
Our goal is to learn a missing-aware representation function
$f$ that maps the multi-modal inputs of $(r_i,t)$ to a $d$-dimensional region embedding:
\begin{equation}
f:\Big(r_i, t, \{\mathbf{x}^{(m)}_{i,t}, \mathbf{s}^{(m)}_{i,t}\}_{m\in\mathcal{M}}\Big)\rightarrow \mathbf{Z}_{i,t}\in\mathbb{R}^{d}.
\end{equation}

The learned embedding $\mathbf{Z}_{i,t}$ should capture region characteristics while being robust to structured missingness, and is used for the downstream house rent prediction task:
\begin{equation}
\hat{y}_{i,t+1}=MLP(\mathbf{Z}_{i,t}),
\end{equation}
where $y_{i,t+1}$ is the ground-truth rent for the next month.

\begin{figure*}
    \centering
    \includegraphics[width=0.9\linewidth]{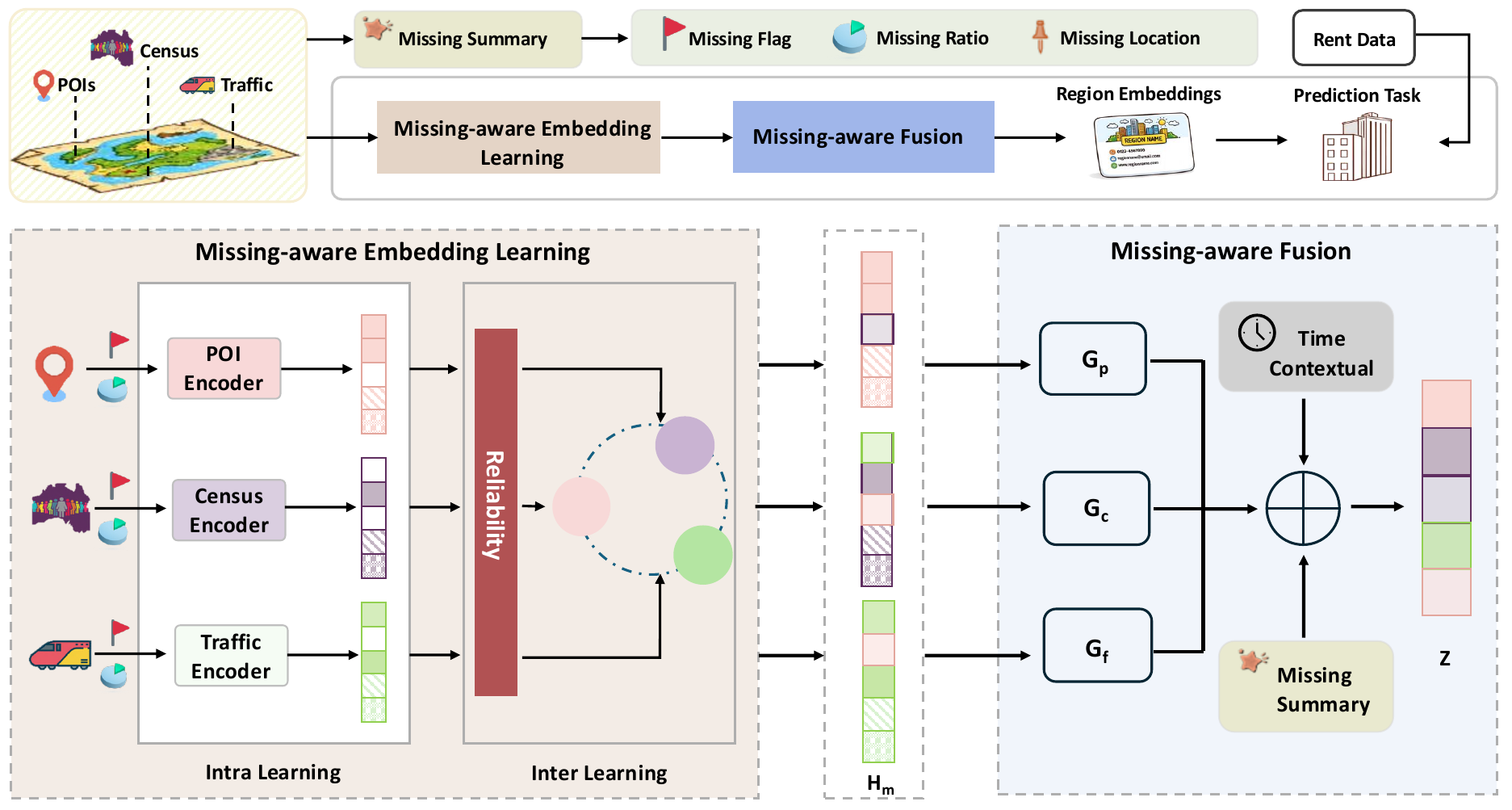}
    \caption{Overview of the Proposed MARCUS Framework}
    \label{fig:model}
\end{figure*}

In this section, we provide a detailed description of our proposed model MARCUS. First, we introduce Missing-aware Embedding Learning, which consists of two branches: Intra Learning and Inter Learning. These two branches jointly produce embeddings that are sensitive to missingness patterns and automatically calibrate modality trustworthiness. Then, we present Missing-aware Fusion, which dynamically weights and fuses modality representations based on missingness patterns and temporal context to obtain more robust embeddings. The overall architecture of the model is illustrated in Fig.~\ref{fig:model}. It mainly consists of two modules. First, we feed the preprocessed multimodal inputs into Missing-aware Embedding Learning. The inputs include modality features and the corresponding missing summary. This module has two stages, Intra Learning and Inter Learning. Intra Learning jointly models observed-value features and missingness features within each modality. It produces a missing-aware representation for each modality. Inter Learning then performs cross-modal interaction. It estimates modality reliability from the missingness status. It also adaptively calibrates the strength of information exchange across modalities. This produces updated cross-modal embeddings. Next, these embeddings are sent to Missing-aware Fusion. Fusion uses the missing summary and the temporal context. It dynamically weights and fuses modality representations. Finally, it outputs the region representation for downstream rent prediction tasks.

\subsection{Missing-aware Embedding Learning}
This section describes our missing-aware embedding learning, where missingness is incorporated into intra-modal encoding and cross-modal interaction.

\subsubsection{Intra Learning: Modality-specific Missing-aware Encoding}

For each region $r_i$ and month $t$, we consider a set of modalities $\mathcal{M}$ (e.g., POI, Census, and Traffic). For each modality $m \in \mathcal{M}$, we construct the modality input by concatenating its content embedding and the missingness feature vector:
\begin{equation}
  \mathbf{x}^{(m)}_{i,t}=\big[\mathbf{e}^{(m)}_{i,t}\ \Vert\ \mathbf{s}^{(m)}_{i,t}\big],
\end{equation}
where $\mathbf{e}^{(m)}_{i,t}\in\mathbb{R}^{d_e^m}$ denotes the modality embedding, and the missingness feature vector is defined as
\begin{equation}
  \mathbf{s}^{(m)}_{i,t}=\big[b^{(m)}_{i,t},\ \rho^{(m)}_{i,t},\ \mathbf{l}^{(m)}_{i,t}\big]\in\mathbb{R}^{d_s^m}.
\end{equation}
Note that POI and Census are static modalities, thus $\mathbf{e}^{(m)}_{i,t}\equiv \mathbf{e}^{(m)}_{i}$ for $m\in\{\text{P},\text{C}\}$. Importantly, Intra Learning does not take the month index $t$ as an explicit input; temporal context is incorporated later in the fusion stage.

We instantiate Intra Learning with a lightweight MLP-based encoder block $f_{\text{enc}}^{(m)}(\cdot)$ for each modality $m$. All modalities share the same encoder architecture while using modality-specific parameters. The encoder maps $(\mathbf{e}^{(m)}_{i,t}, \mathbf{s}^{(m)}_{i,t})$ to a unified hidden representation $\mathbf{h}^{(m)}_{i,t}\in\mathbb{R}^{d_h}$ by 1) projecting the content embedding and the missingness vector into the same hidden space via two separate branches and 2) fusing the two branches through a fusion layer.

Specifically, we first compute two branch-wise hidden vectors:
\begin{align}
\tilde{\mathbf{h}}^{(m)}_{e} &= \sigma\!\left(\mathbf{W}^{(m)}_{e}\mathbf{e}^{(m)}_{i,t}+\mathbf{b}^{(m)}_{e}\right),\\
\tilde{\mathbf{h}}^{(m)}_{s} &= \sigma\!\left(\mathbf{W}^{(m)}_{s}\mathbf{s}^{(m)}_{i,t}+\mathbf{b}^{(m)}_{s}\right).
\end{align}

where $\sigma(\cdot)$ is a non-linear activation function. We then concatenate the two hidden vectors and apply a fusion layer to obtain the final intra-modality representation:
\begin{equation}
  \mathbf{h}^{(m)}_{i,t}=\sigma\!\left(\mathbf{W}^{(m)}_{f}\big[\tilde{\mathbf{h}}^{(m)}_{e}\ \Vert\ \tilde{\mathbf{h}}^{(m)}_{s}\big]+\mathbf{b}^{(m)}_{f}\right).
\end{equation}
The resulting $\mathbf{h}^{(m)}_{i,t}$ serves as the modality-specific hidden representation for subsequent Inter Learning and Missing-aware Fusion.

\subsubsection{Inter Learning: Reliability-aware Cross-modality Interaction}
After Intra Learning, we obtain intra-modality representations $\mathbf{h}^{(m)}_{i,t}\in\mathbb{R}^{d_h}$ for each modality $m\in\mathcal{M}=\{P,C,F\}$. Since these modalities provide complementary urban signals (e.g., amenities, socioeconomic attributes, and mobility patterns), \textit{Inter Learning} aims to explicitly model cross-modality dependencies so that each modality can absorb complementary cues from the others and refine its representation, resulting in inter-updated embeddings $\mathbf{H}^{(m)}_{i,t}$. Meanwhile, real-world observations are often incomplete and exhibit heterogeneous data quality across modalities. Simply performing cross-modality interaction may propagate noise from severely missing modalities. To address this issue, we introduce a reliability-aware interaction mechanism that adaptively suppresses the influence of unreliable modalities during cross-modality interaction.

\textbf{Modality Reliability Estimation.}
We treat missingness as a useful signal reflecting data availability and quality. Given the missingness vector
$\mathbf{s}^{(m)}_{i,t}=\big[b^{(m)}_{i,t},\rho^{(m)}_{i,t},\mathbf{l}^{(m)}_{i,t}\big]$, Inter Learning focuses on the two most direct quality indicators: the missing  flag and the missingness ratio.
Specifically, we estimate a scalar reliability weight $\alpha^{(m)}_{i,t}\in(0,1)$ from $\big[b^{(m)}_{i,t},\rho^{(m)}_{i,t}\big]$ via a lightweight mapping:
\begin{equation}
  \alpha^{(m)}_{i,t}=\mathrm{sigmoid}\!\left(\mathbf{w}_\alpha^\top \big[b^{(m)}_{i,t},\ \rho^{(m)}_{i,t}\big]+b_\alpha\right),\quad m\in\{P,C,F\}.
\end{equation}
Intuitively, $\alpha^{(m)}_{i,t}$ quantifies the trustworthiness of modality $m$ under the current observation pattern. It decreases when the modality is less available or more severely missing, and is therefore used to down-weight its contribution during cross-modal interaction.

\textbf{Reliability-modulated Cross-modality Interaction.}
We model cross-modality interaction by applying multi-head self-attention over the modality dimension, where each modality representation attends to the others to aggregate complementary information. Let $\Delta\mathbf{h}^{(m)}_{i,t}$ denote the interaction-induced update for modality $m$ produced by the cross-modality attention module. To prevent unreliable modalities from amplifying noisy interaction effects, we modulate the update magnitude using the estimated reliability and fuse it with a residual connection:
\begin{equation}
  \mathbf{H}^{(m)}_{i,t}=\mathbf{h}^{(m)}_{i,t}+\alpha^{(m)}_{i,t}\cdot \Delta\mathbf{h}^{(m)}_{i,t},\quad m\in\{P,C,F\}.
\end{equation}
In this way, modalities with higher reliability preserve more interaction updates and benefit more from complementary cues, while unreliable modalities are explicitly suppressed to reduce noise propagation. In practice, normalization and a feed-forward network can be further applied for training stability and expressive power. However, the core idea remains that missingness-driven reliability adaptively controls the strength of cross-modality interaction.

The resulting $\{\mathbf{H}^{(m)}_{i,t}\}_{m\in\mathcal{M}}$ are then fed into the subsequent Missing-aware Fusion module to produce the final region embedding.

\subsection{Missing-aware Fusion: Dynamic Fusion with Missingness and Temporal Context}
After Inter Learning, we obtain inter-updated modality representations $\mathbf{H}^{(m)}_{i,t}\in\mathbb{R}^{d_h}$ for each modality $m\in\mathcal{M}=\{P,C,F\}$. Since the contribution of each modality may vary across months (for example,  traffic dynamics) and observations are often incomplete with heterogeneous missingness patterns, a fixed fusion strategy (such as uniform averaging) can be suboptimal and unstable. We therefore propose \textit{Missing-aware Fusion}, which leverages both missingness patterns and temporal context to dynamically assign modality weights and produce a robust region representation $\mathbf{Z}_{i,t}$ for downstream prediction.

\textbf{Temporal Context Encoding.}
We map the month index $t$ to a temporal context vector $\mathbf{c}_t$ via a learnable embedding followed by a projection. Importantly, this temporal signal is not directly used in Intra or Inter encoding. Instead, it is introduced at the fusion stage to capture month-specific modality contributions and seasonal effects.

\textbf{Missingness and Time-conditioned Gating.}
For each modality $m$, we compute a scalar fusion weight $\lambda^{(m)}_{i,t}\in(0,1)$ using a modality-specific gating function (a lightweight MLP) conditioned on the full missingness vector and temporal context:
\begin{equation}
  \lambda^{(m)}_{i,t}=g_m\!\left(\big[\mathbf{s}^{(m)}_{i,t}\ \Vert\ \mathbf{c}_t\big]\right),\quad m\in\{P,C,F\}.
\end{equation}
Compared to Inter Learning, which estimates reliability using only $\big[b^{(m)}_{i,t},\rho^{(m)}_{i,t}\big]$, Missing-aware Fusion exploits the complete missingness vector
$\mathbf{s}^{(m)}_{i,t}=\big[b^{(m)}_{i,t},\rho^{(m)}_{i,t},\mathbf{l}^{(m)}_{i,t}\big]$ to capture where missingness occurs and its spatial pattern, enabling finer-grained and context-dependent weighting.

\textbf{Normalized Weighted Aggregation.}
Given the dynamic weights, we fuse modality representations via a normalized weighted sum:
\begin{equation}
  \mathbf{Z}_{i,t}
  =
  \frac{
    \sum_{m\in\mathcal{M}} \lambda_{i,t}^{(m)} \mathbf{H}_{i,t}^{(m)}
  }{
    \sum_{m\in\mathcal{M}} \lambda_{i,t}^{(m)} + \epsilon
  }.
\end{equation}
where $\epsilon$ is a small constant for numerical stability. This design allows the model to down-weight modalities that are severely missing or less informative under the current month, and to emphasize more reliable and month-relevant modalities, resulting in robust region embeddings $\mathbf{Z}_{i,t}$ for downstream tasks like rent prediction.

\section{Experiments}

Our experiments are designed to address four research objectives:

\textbf{1)} Performance comparison with traditional machine learning methods and existing region-embedding approaches.

\textbf{2)} Applicability of MARCUS across different countries and geographic scales;

\textbf{3)} Effectiveness of region embeddings for rent prediction tasks with pronounced seasonal fluctuations, as shown in Fig.~\ref{fig:rent}(b);

\textbf{4)} Contribution of each module to overall performance (via ablation studies).

\textbf{Dataset.} We conduct experiments on real-world datasets from two cities: Sydney and New York. We collect region partitions, census data, POI data, traffic records, and rent data. For region partitioning, we use SA2 area for Sydney and ZIP codes area (postcodes) for New York. More details on the data description and analysis can be found in Section 2.

\textbf{Baselines.} We compare MARCUS with two categories of baselines. The first category includes traditional machine learning predictors: LR, SVR, and RF. These baselines are used to examine whether incorporating region embeddings improves rent prediction performance. The second category includes existing region embedding models which are used to evaluate whether the representations learned by MARCUS outperform prior methods.They are MVURE \cite{b17}, MGFN \cite{b18}, HREP \cite{b19} and FLEXIREG \cite{b20}.

\textbf{FLEXIREG} proposes learning more flexible representations of urban regions. It learns cell-level embeddings on a hexagonal grid from multimodal data, including POIs, street-view images and land-use data. Then, FLEXIREG customizes the representations for different region partitions and downstream tasks via adaptive aggregation and prompt learning.

\textbf{HREP} targets heterogeneous region embedding. The core innovation lies in introducing prompt learning to region embedding. Task-specific prompts guide downstream task learning, thereby eliminating the need to retrain region representations for each individual task.

\textbf{MGFN} uses human mobility as the sole modality to mine urban structure and patterns. It fuses mobility graphs from multiple time steps into mobility patterns, and learns region representations via intra-pattern message passing combined with inter-pattern cross-attention.

\textbf{MVURE} uses multi-view urban data, including mobility views and urban attribute views. It encodes each view with GAT, shares information via cross-view self-attention, and fuses the views with adaptive weights into a region embedding.

We similarly learn region embeddings by adaptively fusing multimodal urban data. MARCUS further addresses coverage imbalance and missingness by treating them as reliability signals that guide cross-modal interaction and fusion, thereby enabling reliable downstream performance even when certain features are missing.

We also include a zero-information baseline (\textbf{ZIB}), which predicts the training-set mean of the target for every test sample without using any input features.It serves as a two-sided reference: compared with ZIB, lower error indicates retained region-discriminative information, while higher error suggests that the model is misled by its inputs under severe missingness.

\textbf{Metrics.} For the downstream task, we use three evaluation metrics to assess prediction performance: mean absolute error (MAE), root mean square error (RMSE), and mean absolute percentage error (MAPE). Lower values indicate better performance.

\subsection{Overall Performance}

Table~3 reports the overall results of MARCUS and all baselines across the three metrics. MARCUS achieves state-of-the-art performance on all metrics. For clarity, the best overall result is highlighted in \textbf{bold}, while the best-performing baseline is marked with \underline{underlining}. Based on these results, we have the following findings:

{\setlength{\tabcolsep}{1.6pt} 
\begin{table}[t]
  \centering
  \caption{Performance comparison on Sydney and New York}
  \label{tab:resul}
  \begin{tabular}{l c c c c c c}
    \toprule
    \multirow{2}{*}{Model} &
    \multicolumn{3}{c}{Sydney} &
    \multicolumn{3}{c}{New York} \\
    \cmidrule(lr){2-4} \cmidrule(lr){5-7}
    & MAE $\downarrow$ & RMSE $\downarrow$ & MAPE(\%) $\downarrow$ & MAE $\downarrow$ & RMSE $\downarrow$ & MAPE(\%) $\downarrow$ \\
    \midrule
    ZIB &
    126.46 & 158.65 & 24.05 &
    558.94 & 862.07 & 16.14 \\
    \midrule
    LR   & 112.59 & 142.80 & 21.69 & 353.18 & 511.62 & 10.92 \\
    SVR  &  94.40 & 105.73 & 16.72 & 336.75 & 381.17 & 11.14 \\
    RF   &  84.46 &  96.61 & 15.19 & 187.42 & 261.18 &  6.04 \\
    \midrule
    FLEXIREG &  77.18 &  88.83 & 14.18 & \underline{160.10} & \underline{239.55} &  \underline{5.14} \\
    HREP &  75.51  &  87.33 & 13.97 & 174.81 & 251.69 &  5.61 \\
    MGFN &  90.28 & 102.25 & 16.36 & 196.41 & 276.09 &  6.42 \\
    MVURE&  \underline{72.54} &  \underline{85.87} & \underline{13.45} & 189.48 & 287.29 &  6.17 \\
    \midrule
    \textbf{MARCUS} &
    \textbf{35.29} & \textbf{45.91} & \textbf{7.08} &
    \textbf{139.90} & \textbf{213.87} & \textbf{4.51} \\
    Enhance(\%) & 51.35 & 46.53 & 47.36 & 12.62 & 10.72 & 12.26 \\
    \bottomrule
  \end{tabular}
\end{table}
}

\textbf{1)} MARCUS achieves strong overall performance and outperforms all baseline models. In terms of MAE, MARCUS improves by up to 51.35\% over the best baseline MVURE in the Sydney rental market. It also consistently surpasses the previous SOTA baseline FLEXIREG across all evaluated metrics and datasets, achieving improvements of 54.2\% in Sydney and 12.62\% in New York. The larger gain in Sydney reflects baseline degradation under native missingness rather than instability of MARCUS. Relative to ZIB, MARCUS stays at 25--29\% of the zero-information error in both cities, whereas the strongest competing model degrades from 28--32\% on New York to 54--57\% on Sydney. This is largely due to Sydney's data, where 40.4\% of traffic records are natively missing, a regime that baseline models without explicit missingness handling struggle to handle.


\textbf{2)} MARCUS remains effective under different geographic partition schemes. As mentioned earlier, Sydney follows a statistical geography standard, while New York uses postal zones. The consistent performance across these two different rental market settings and spatial scales suggests that MARCUS has good applicability beyond a single city-specific partition scheme.


\textbf{3)} MARCUS is effective for rent forecasting with temporal dynamics. Since rental markets exhibit pronounced seasonal fluctuations, we incorporate time into downstream prediction by appending the same time embedding to all models. We do not modify the baseline representation learning models, ensuring a fair and consistent use of temporal information. Under this setting, MARCUS still outperforms all baselines, further demonstrating its effectiveness for seasonal rent forecasting.


\subsection{Ablation Study}
We conduct the following studies to validate the effectiveness of the components in MARCUS: 

\textbf{1) MARCUS-w/o-IAM}: We disable the use of missingness features in the Intra Learning module. This investigates whether missingness should be incorporated early during each modality’s representation learning, or whether it is sufficient to introduce it only in the subsequent Inter learning and Fusion stages. 

\textbf{2) MARCUS-w/o-RW}: We remove the reliability weights computed from the missing flag and missing ratio, thereby treating all modalities equally during cross-modal interactions in the Inter Learning module. It tests whether the gains are mainly due to missingness-based reliability weighting, instead of the cross-modal attention module itself. 

\textbf{3) MARCUS-w/o-FG}: We remove the missing-aware fusion gate in the Fusion stage.It examines whether the final representation benefits from missing-aware fusion weighting, or whether naive fusion is already sufficiently robust. 

\textbf{4) MARCUS-w/o-TG}: We keep the gating structure but remove the time context input, computing fusion weights solely from missingness features. This variant assesses whether time-conditioned gating is necessary for rent prediction tasks with trends and seasonality. 

\textbf{5) MARCUS-w/o-M}: We completely remove missingness from the model. This control experiment is designed to quantify the benefit of treating missingness as an urban contextual signal.

\begin{figure}[t]
    \centering
    \includegraphics[width=1\linewidth]{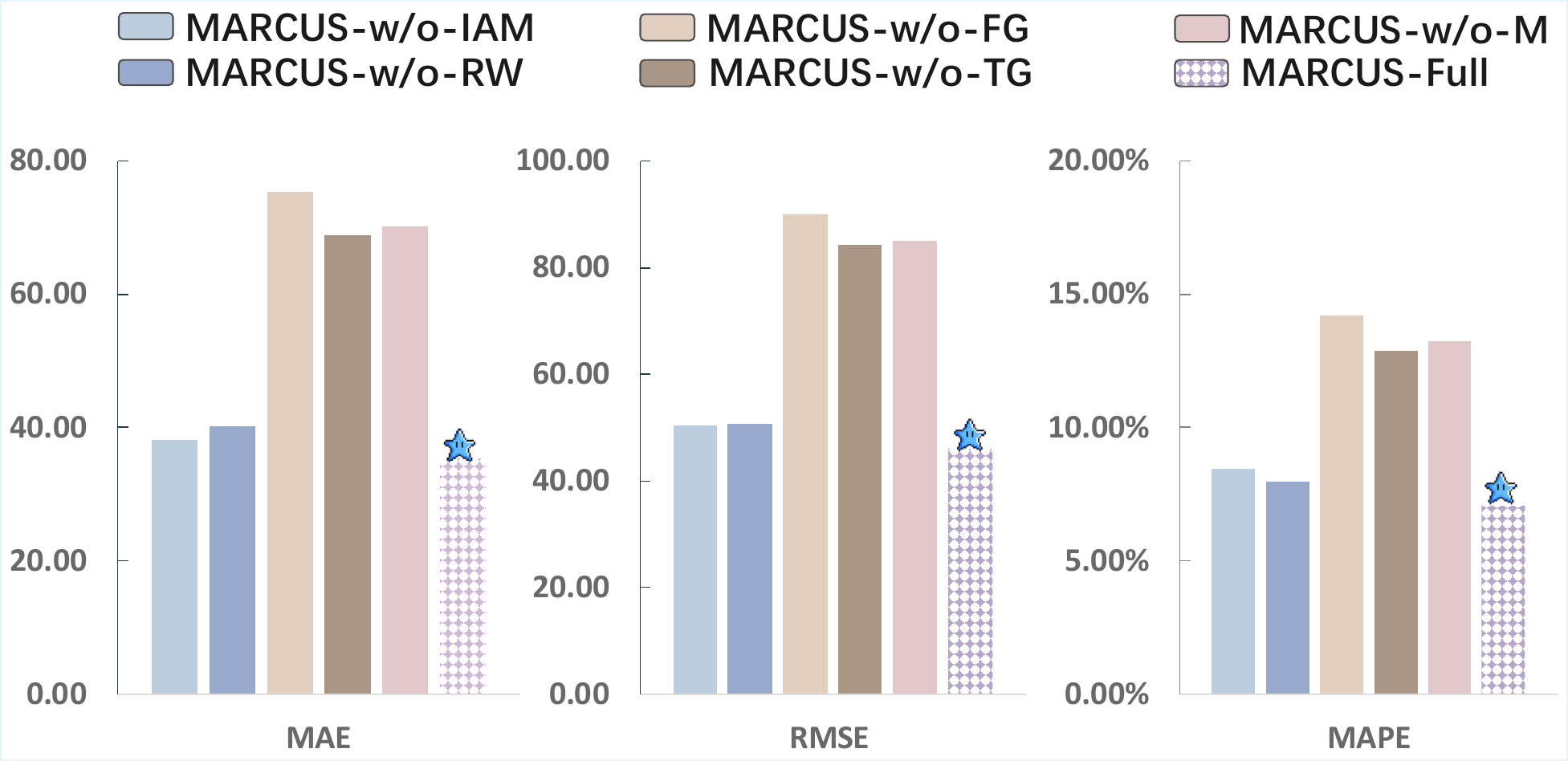}
    \caption{Ablation Study Results (Sydney)}
    \label{fig:ablation}
\end{figure}

As shown in Fig.~\ref{fig:ablation}, MARCUS consistently outperforms all variants, highlighting the contribution of each component to its overall effectiveness. The main observation is that missingness should be treated as a useful contextual signal, rather than being handled only as a data quality issue. When missingness is completely removed (\textbf{MARCUS-w/o-M}), the prediction performance degrades drastically that MAE nearly doubles from 35.29 to 70.15. Second, the missing-aware fusion gate and its temporal conditioning constitute the key contributors to performance. Removing the fusion gate (\textbf{MARCUS-w/o-FG}) results in the largest performance drop. This implies that missingness should inform the fusion weights at the final stage, as naive fusion may override earlier missing-aware benefits. Moreover, removing time context from the gate (\textbf{MARCUS-w/o-TG}) leads to a substantial degradation. Because in rent prediction, long-term trends and seasonality make missingness effects time-dependent. The same missingness pattern can reflect different semantics and modality reliability across months, motivating temporal conditioning in the fusion gate.

After confirming the decisive role of the fusion stage, we further examine the effect of missingness in earlier stages, including reliability modulation in cross-modal interactions (\textbf{MARCUS-w/o-RW}) and intra-modality missing-aware encoding (\textbf{MARCUS-w/o-IAM}). Specifically, without reliability weighting, MAE increases from 35.29 to 40.24, indicating that missingness-driven reliability modeling effectively suppresses noise propagation in cross-modal attention.  In contrast, removing intra-modality missing-aware encoding results in a smaller degradation. Overall, these observations suggest that missingness matters most in interaction and fusion, whereas intra-modality missing-aware encoding offers complementary gains in alignment and robustness. The ablation results are detailed in Table 4.

\begin{table}[t]
  \caption{Performance comparison of MARCUS variants.}
  \label{tab:MARCUS_ablation}
  \centering
  \begin{tabular}{lccc}
    \toprule
    Model & MAE & RMSE & MAPE(\%) \\
    \midrule
    \textbf{MARCUS-w/o-IAM} & 38.24 & 50.34 & 8.48 \\
    \textbf{MARCUS-w/o-RW}  & 40.24 & 50.63 & 7.96 \\
    \textbf{MARCUS-w/o-FG}  & 75.39 & 90.08 & 14.20 \\
    \textbf{MARCUS-w/o-TG}  & 68.86 & 84.27 & 12.87 \\
    \textbf{MARCUS-w/o-M}   & 70.15 & 85.09 & 13.24 \\
    \midrule
    \textbf{MARCUS-Full} & \textbf{35.29} & \textbf{45.91} & \textbf{7.08} \\
    \bottomrule
  \end{tabular}
\end{table}

\subsection{Analysis on Missing Ratio}

We further conduct additional experiments with different levels of additional missingness on the rental price prediction task. Specifically, the original missing data are kept unchanged, while a certain proportion of originally observed data is randomly masked. 

Fig.~\ref{fig:ratio} compares MARCUS with baseline models on the Sydney dataset. As the additional missing ratio increases, MARCUS degrades smoothly and monotonically. When all originally observed modality cells are masked, its MAE reaches 116.2, which remains below ZIB. This indicates that even when all originally observed modality information is removed, MARCUS can fall back on a meaningful learned prior through explicit missingness encoding, rather than being misled by absent inputs.

In contrast, the baselines fail in distinct ways. MVURE's MAE jumps by 55\% from 71.4 to 110.5 with only 5\% additional masking, and then fluctuates around ZIB. FlexiReg crosses ZIB at a ratio of around 0.3 and plateaus above it, suggesting limited region-discriminative information. MGFN deteriorates monotonically to 176.8, about 40\% worse than ZIB, indicating that masked inputs may actively mislead it. HREP appears robust but is mainly insensitive to attribute modalities. Its MAE without additional masking is already 2.3$\times$ that of MARCUS (81.4 versus 35.3), so it has little modality information to lose when originally observed modality data are removed.

Overall, MARCUS outperforms all baselines throughout the practically relevant regime, where the additional missing ratio is no larger than 0.5, while being the only model that both fully exploits available modalities and remains below ZIB under severe missingness.

\section{Real-world Deployment}
\begin{figure}
    \centering
    \includegraphics[width=0.8\linewidth]{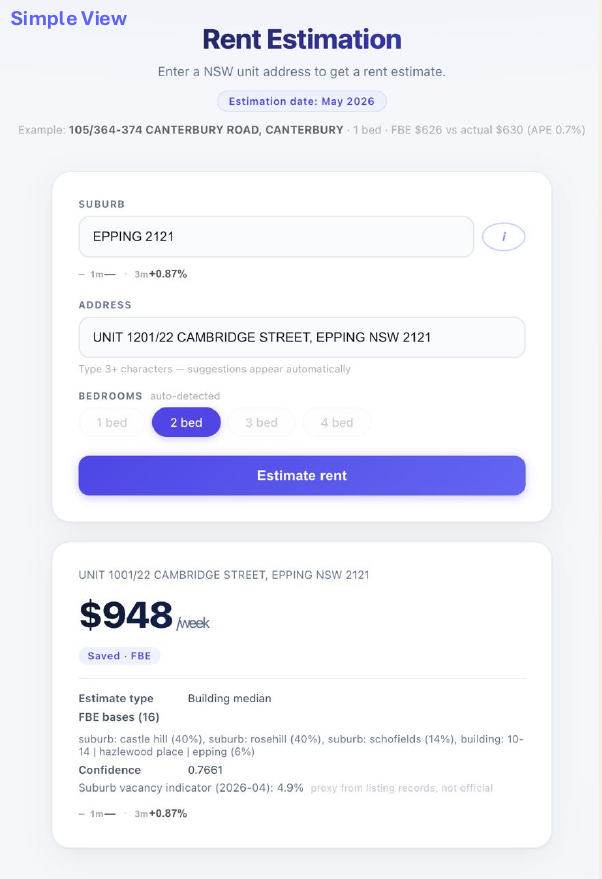}
    \caption{Sydney Rent Prediction System-Simple View}
    \label{fig:demo}
\end{figure}

\begin{figure}
    \centering
    \includegraphics[width=1\linewidth]{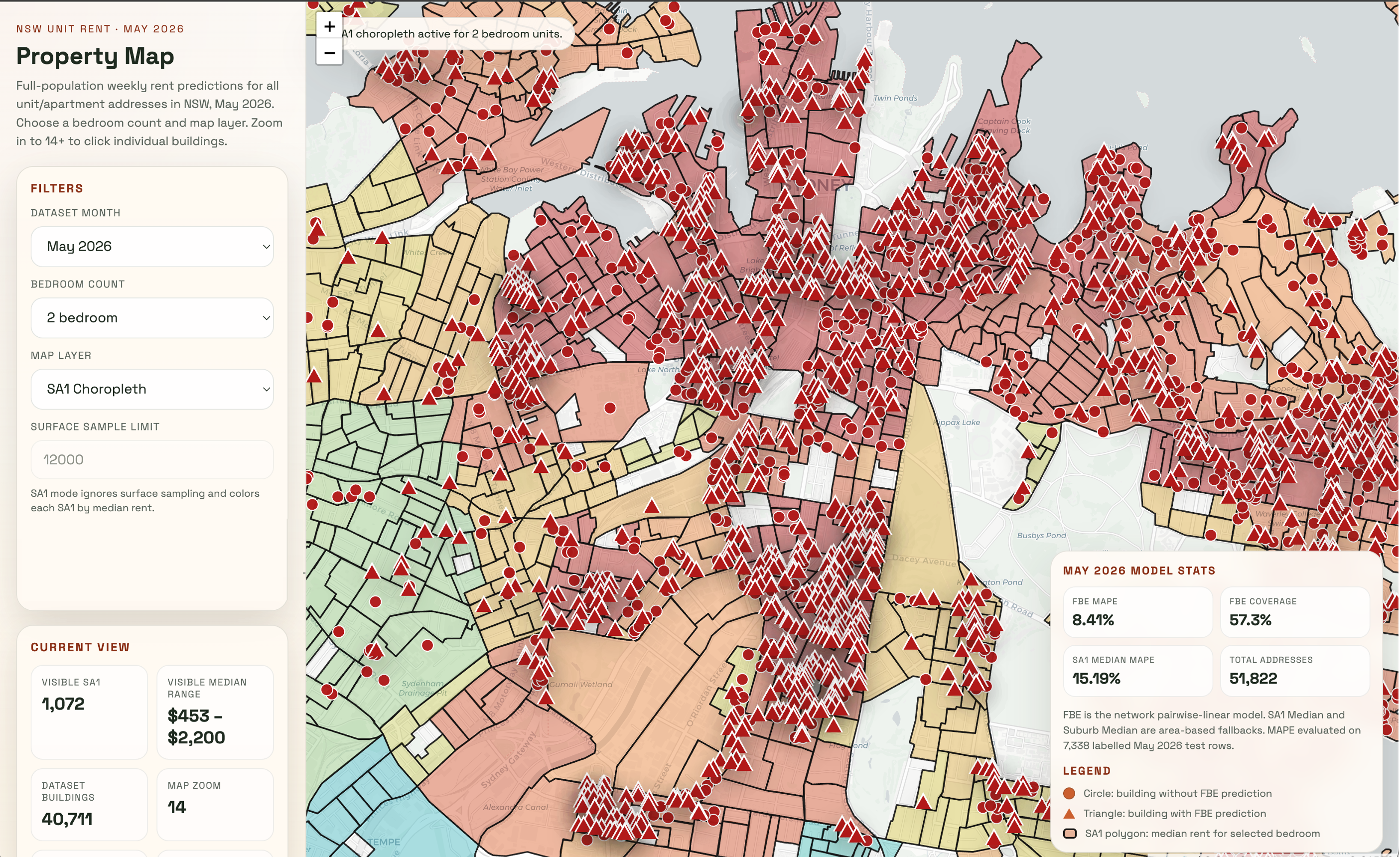}
    \caption{Sydney Rent Prediction System-Professional View}
    \label{fig:demo2}
\end{figure}

The proposed MARCUS model has been deployed in a real-world rent prediction system operated by our industry partner, Property Investors Alliance (PIA), as shown in Fig.~\ref{fig:demo} and Fig.~\ref{fig:demo2}. The system provides two modes: a simple view and a professional view. The simple view supports basic rent prediction, while the professional view extends it with additional functions, including file upload and map-based location.


PIA is an Australian residential property investment and property management platform that provides data-driven market analysis and advisory services, and manages approximately 6,000 properties. In the deployed system, MARCUS generates region-level rent estimates across Greater Sydney, with Statistical Area Level 2 (SA2) regions serving as the default spatial unit. These estimates are combined with predictions at other spatial scales (SA1 and suburb) to provide comprehensive rental recommendations.
The online deployment incorporates continuously updated multimodal data beyond the temporal range used in offline experiments, including newly collected rental transactions and urban signals from July 2024 onward, reflecting realistic operational conditions with persistent and heterogeneous missingness.

Under this real-world deployment, MARCUS achieves a 57.1\% improvement in rent prediction accuracy at the SA2 level compared with PIA's previously adopted model, ST-ResMAE~\cite{b6}. Beyond accuracy gains, the deployment demonstrates clear practical benefits in reducing uncertainty in region-level rent estimation under incomplete multimodal observations. By producing more stable and informative regional rent reference ranges, the system supports downstream pricing and investment decision-making while avoiding reliance on individual-level or sensitive personal data.

An additional practical advantage of MARCUS is its flexibility with respect to spatial granularity. Although SA2 regions are used by default due to their alignment with official statistical reporting, the model can be readily adapted to finer-grained units such as SA1 regions or to suburb-level partitions that are more intuitive for end users. This adaptability allows the deployed system to support diverse consumer-facing, business, and policy-oriented urban analytics, demonstrating that explicitly modelling missingness delivers measurable practical value beyond offline benchmarks.

\section{Conclusion}

In this paper, we studied urban region representation learning under incomplete multimodal urban data, where systematic missingness is prevalent and often correlated with regional attributes and urban structure. Rather than treating missingness as noise to be imputed or masked, we propose MARCUS, which models missingness patterns as contextual urban signals to guide region representation learning and multimodal fusion. Specifically, MARCUS incorporates missingness into intra-modality encoding and reliability-aware cross-modality interaction. It further performs missingness and time-conditioned fusion to learn region embeddings that are robust to incompleteness and semantically informative. Experiments on real-world datasets from Sydney and New York for rent prediction show that MARCUS consistently outperforms all baselines. In addition, we include an ablation setting that imputes missing values and disables all missing-aware components. Performance drops noticeably under this setting, confirming that explicitly modelling missingness is effective. Overall, these results highlight the practical value of missing signals in multimodal urban computing.

\end{document}